\documentclass{article}
\usepackage{jfrExamplee}
\usepackage{graphicx}
\usepackage{apalike}
\usepackage{setspace}
\usepackage{graphicx,soul,framed,mathptmx,times,amsmath,amssymb,color,
flushend,gensymb,subfigure,fancyhdr,subfigure,lipsum,amsthm, amsfonts}
\usepackage{cite,siunitx}

\DeclareMathOperator{\atantwo}{atan2}

\usepackage{framed}
\definecolor{shadecolor}{rgb}{1,0.8,0.3}

\usepackage[normalem]{ulem}

\allowdisplaybreaks

\title{High Precision Control of Tracked Field Robots in the Presence of Unknown Traction Coefficients}

\author{
Erkan~Kayacan \\
Distributed Autonomous Systems Laboratory \\
Coordinated Science Laboratory   \\
University of Illinois at Urbana-Champaign\\
Urbana, IL 61801 \\
\texttt{erkank@mit.edu} 
\And
Sierra~N. ~Young \\
Department of  Civil and Environmental Engineering \\
University of Illinois at Urbana-Champaign\\
Urbana, IL 61801 \\
\texttt{snyoung2@illinois.edu} 
\AND
Joshua~M. ~Peschel  \\
Dept. of Agricultural and Biosystems Engineering \\
 Iowa State University\\
 Ames, Iowa 50011-3270 \\
\texttt{ peschel@iastate.edu} 
\And
Girish~Chowdhary \\
Distributed Autonomous Systems Laboratory \\
Coordinated Science Laboratory   \\
Dept. of Agricultural and Biological Engineering \\
University of Illinois at Urbana-Champaign\\
Urbana, IL 61801 \\
\texttt{girishc@illinois.edu} 
}

\begin{document}

\maketitle

\begin{abstract}

Accurate steering through crop rows that avoids crop damage is one of the most important tasks for agricultural robots utilized in various field operations, such as monitoring, mechanical weeding, or spraying.  In practice, varying soil conditions can result in off-track navigation due to unknown traction coefficients so that it can cause crop damage. To address this problem, this paper presents the development, application, and experimental results of a real-time receding horizon estimation and control (RHEC) framework applied to a fully autonomous mobile robotic platform to increase its steering accuracy. Recent advances in cheap and fast microprocessors, as well as advances in solution methods for nonlinear optimization problems, have made nonlinear receding horizon control (RHC) and receding horizon estimation (RHE) methods suitable for field robots that require high frequency (milliseconds) updates. A real-time RHEC framework is developed and applied to a fully autonomous mobile robotic platform designed by the authors for in-field phenotyping applications in Sorghum fields. Nonlinear RHE is used to estimate constrained states and parameters, and nonlinear RHC is designed based on an adaptive system model which contains time-varying parameters. The capabilities of the real-time RHEC framework are verified experimentally, and the results show an accurate tracking performance on a bumpy and wet soil field. The mean values of the Euclidean error and required computation time of the RHEC framework are respectively equal to $0.0423$ m and $0.88$ milliseconds. 
\end{abstract}

\section{Introduction}\label{sec_intro}

Due to the petroleum crisis and growing demands for renewable energy sources \cite{Tverberg2012}, there have been increased research efforts to improve bioenergy crop breeding for biofuel production \cite{Wang2016,buckeridge2014plants}. The success and sustainability of biofuels are dependent on increasing plant yields  \cite{Vega2010, Yano2009}. Top-yielding plant traits (phenotypes) are linked to their corresponding genes (genotypes). Identifying useful phenotypes can help in identifying which genes will increase the productivity of biofuel crops. However, the process of phenotyping is cost and labor intensive. On the other hand, stationary high throughput phenotyping platforms, or greenhouse based phenotyping systems are bulky, costly, and require elaborate infrastructure \cite{Fiorani2013}. Therefore, there is a high need for an automated, mobile, sensory platform that could significantly increase the throughput and accuracy of screening crops in the field.

\begin{figure}[h!]
  \centering
  \includegraphics[width=3.4in]{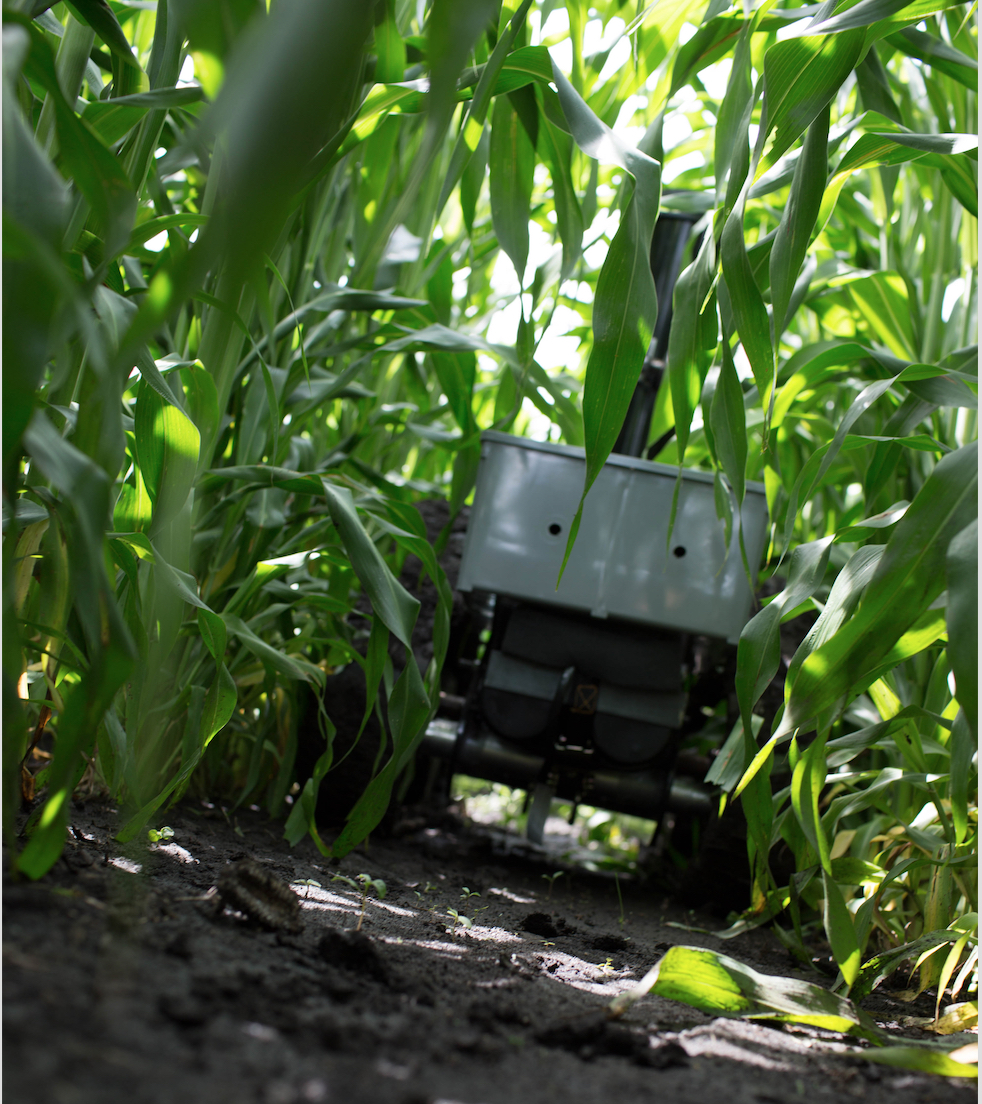}\\
  \caption{Field robot in sorhgum plants in Energy Farm, Urbana, IL, USA.}\label{fig_robotsorghum}
\end{figure}

One of the major challenges in developing robotic phenotyping platforms is high-precision guidance and control in muddy and uneven agricultural fields \cite{KAYACAN2014926, 6695753}. A high-precision control scheme is necessary to avoid crop disturbance and damage. For example, Figure \ref{fig_robotsorghum} depicts the TERRA-MEPP phenotyping robot under development at the University of Illinois at Urbana Champaign. It is a typical tracked robot that carries hyperspectral, infrared, visual spectrum, and lidar sensors for phenotyping. Due to the necessity to carry large sensing payloads, the width a phenotyping robot can be significant, for example, it is $0.48$ m for the robot shown in this study. However,  the row spacing of sorghum cultivated for biofuel production may vary from $0.5$ m to $1$ m, with a typical row spacing measured centerline to centerline;  of $0.78$ m as used in this study.  Furthermore, the stem width of a fully mature sorghum plant and the accuracy of the global navigation satellite system used in this study can be up to $0.03$ m so that the effective row spacing is reduced to $0.72$ m, which leaves only $0.12$ m (4.5 in) of space on either side of the robot. Uncertain traction coefficients and uneven fields can make it difficult for the robot to ensure that it stays within the $0.12$ m tolerance so as to not damage the crop. It is the goal of this paper to describe the development of high precision guidance and motion control system that can ensure a navigational tolerance of strictly less than $0.12$ m in real-world field environments.

Tracked robots are preferred to wheeled robots for field-based phenotyping applications due to their improved traction and large contact area with the ground, which minimizes adverse impacts on the soil \cite{Bekker1956}. Tracked robots, however, have complex track-ground interactions and slippage due to differential velocities between treads, which can prompt difficulties in control. The presence of these time-varying parameters can degrade the path-tracking performance when traditional control approaches (e.g., proportional-integral-derivative control) are employed \cite{880812}. The reason is that traditional controllers are not aware of soil parameters of unknown terrain which play a vital role in determining field robot's speed and steering, which are in turn utilized for developing traction control algorithms \cite{KAYACAN2012863}. Therefore, there is a need to develop advanced control algorithms with online parameter estimation to update key terrain and slip parameters, which can be used to improve robot mobility and localization.

Field robots are multi-input-multi-output systems and must be aware of actuator limitations \cite{kayacan2012, 6870427, KAYACAN2016265, 7934317}. Therefore, real-time applications of linear model predictive controllers (MPCs), i.e., RHCs, have been used for path tracking of mobile robots, but there are significant limitations of these linear control techniques \cite{KAYACAN201578}. Early linear MPCs have been designed based on a tracking-error based linear model, and the total control input consists of the feedback and feedforward control actions \cite{Lee2001,Dongbing2006,Barreto2014}. In this scheme, the linear MPC is the feedback controller, and the feedforward control action is derived by taking the reference trajectory and system model into consideration \cite{Klancar2007}. However, these linear control techniques are unable to ensure accurate tracking performance if the systems start off-track or there exist uncertainties, e.g., unknown terrain parameters. Linear control techniques perform well for systems that work at fixed operating-points around which the system is linearized. However, mobile robots work at varying operating points (e.g., tracking of a curvilinear trajectory) and are subjected to several uncertainties (e.g., bumpy fields, variable soil conditions) so that locally linearized models are infeasible. Additionally, linear MPCs cause large system errors in the prediction horizon with a possible instability of the system because the mismatch between the linearized model and the mobile robot increases when mobile robots move away from the target path \cite{Falcone2007}. To overcome these limitations, a robust tube-based MPC has been developed and tested in real-time \cite{7302059}. Although the system could stay-on track, a highly accurate trajectory tracking performance has not been obtained.

To achieve highly accurate tracking performance for constrained navigation in row crops, nonlinear RHC approaches have been widely used, particularly in agricultural applications as they are capable of taking constraints on states and inputs into account and can be designed for nonlinear models \cite{Backman2012, 7125124, Kayacan2018}. Additionally, an accurate online estimation of traction parameters in a system model is crucial as the tire-soil interactions change throughout operations in varying, slippery conditions. Thus, the system model can represent all the system dynamics and interactions in the real-time system. Although nonlinear RHC approach is computationally intensive, availability of compact and powerful computational packages and advances in numerical methods for solving optimization problems have allowed for real-time applications of constrained nonlinear optimization problems for complex systems \cite{7384453,7126158}. Motivated by the previous studies, this study employs a nonlinear receding horizon estimation and control (RHEC) framework for path tracking with a mobile, phenotyping robot.

The main contributions of this study are as follows. 
\begin{itemize}
\item In this work, we augment the kinematic model with traction parameters to capture soil characteristics of agricultural fields, while the system model is specified a priori and remains unchanged during operation in traditional implementations.
\item We provide robust tracking performance when uncertainty is high as uncertainty is reduced through learning whereas traditional RHC approaches do not typically account for model uncertainty. 
\item Highly tracking accuracy, e.g., less than $0.12$ m, for a tracked field robot under unknown and variable soil conditions, has been obtained in real-time by developing and implementing a nonlinear RHEC framework with a sampling time in the range of milliseconds.
\item Computation times for the nonlinear RHEC have been decreased by restricting the number of Gauss-Newton iterations to 1 while solving the nonlinear optimization problems.
\end{itemize}

This paper is organized as follows: Section \ref{sec_mobilerobots} provides the system model with traction parameters and the description of the tracked mobile robot used in this study. Section \ref{sec_rhec} is devoted to describing the nonlinear RHEC approaches. Implementations and solution methods are presented in Section \ref{sec_implementations}. Validation and field tests for the real-time nonlinear RHEC algorithms are presented in Section \ref{sec_experimentalresults}. Finally, conclusions are summarized in Section  \ref{sec_conc}.

\section{Field Robot}\label{sec_mobilerobots}
\subsection{System Description}

\begin{figure}[h!]
  \centering
  \includegraphics[width=3.4in]{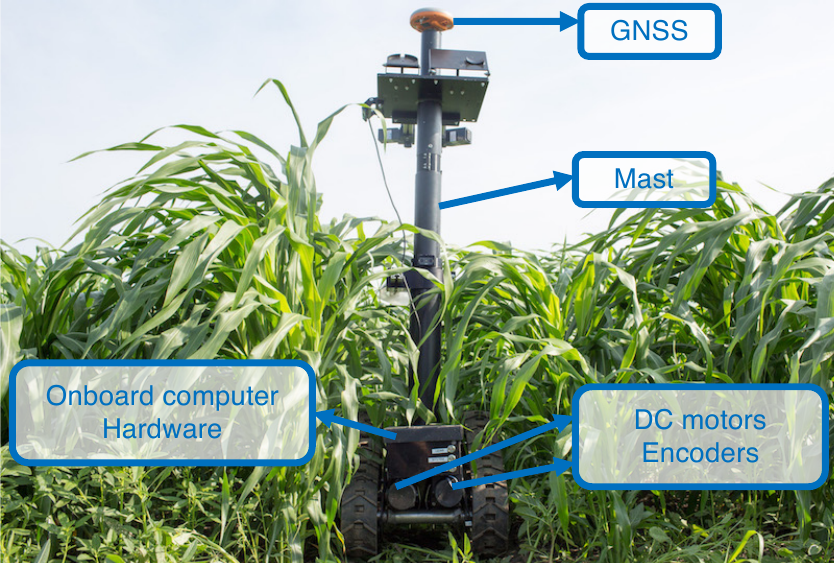}
  \caption{Field robot in sorghum breeding plots in Energy farm, IL, USA. }\label{fig_robot_labeled}
\end{figure}

The field robot as shown in Fig. \ref{fig_robot_labeled} has been built utilizing practical, hands-on experience with modeling, control and estimation techniques, and various sensors and actuators. It is $0.48$ m wide, $1.02$ m long and around $45$ kgs, allowing it to navigate $0.78$ m wide crop rows without damaging the plants and maintaining vertical stability. The mast on the robot can be adjusted to match the sorghum canopy height and is used as a mounting platform for both the plant imaging sensors and a Real-Time Kinematic (RTK) differential Global Navigation Satellite System (GNSS) to acquire positional information. A Septentrio Altus APS-NR2 GNSS receiver (Septentrio Satellite Navigation NV, Belgium) is used to obtain highly accurate positional information, which has a specified position accuracy of $0.03$ m at a $5$-Hz measurements rate. The Trimble network supplies the RTK correction signals via the 4G internet. Two DG-158 A DC motors (SuperDroid Robots Inc., U.S.) capable of $200$ W power output with 16.95 Newton-meter rated torque are used to drive the tracks. E2 optical encoders (US Digital, USA) on DC motors provide speed information with an accuracy of $0.035$ $m/s$. Two 12 V, 40 Ah 4s 20c lipo batteries power the field robot and enable over 3 hours of operation. 

\begin{figure}[h!]
  \centering
  \includegraphics[width=2.8in]{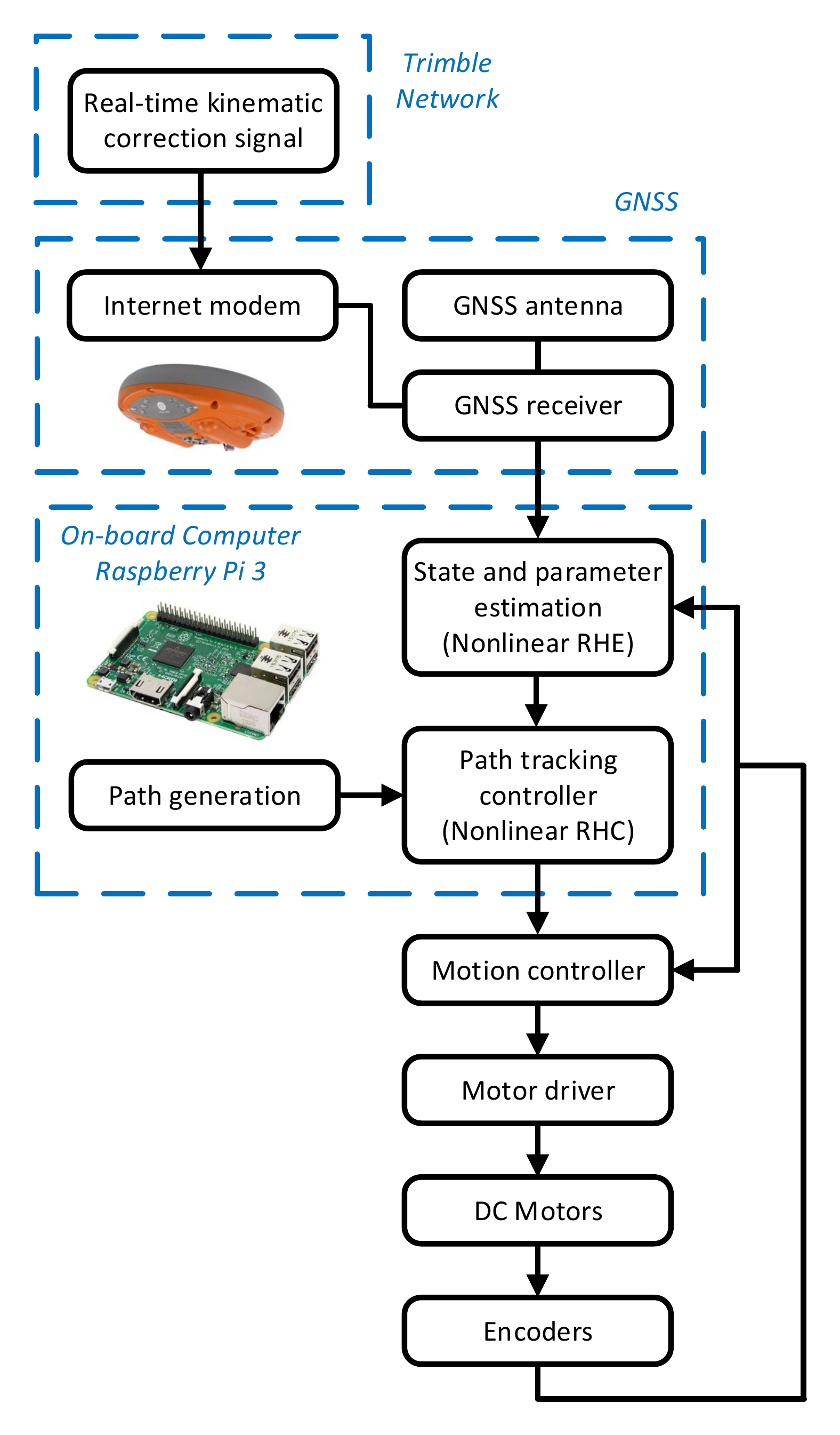}
  \caption{Block diagram of hardware}\label{fig_hardware}
\end{figure}

The real-time estimation and control framework is implemented and executed on an on-board computer (Raspberry Pi 3 Model B), which is equipped with $1.2$ GHz Quad Core Cortex-A53 64-Bit CPU, $1$ GB RAM. It acquires the GNSS data, the speed of the wheels, and controls the tracked robot by sending signals to the Kangaroo x2 motion controller (Dimension Engineering, USA). The generated control input is sent to the motion controller via the on-board computer, and speed measurements of the wheels are received with the same way. Positional measurements are obtained via Bluetooth. The sampling frequency is set to $5$-Hz due to the fact that the maximum update rate of the GNSS is equal to $5$-Hz. 

The block diagram of the hardware is illustrated in Fig. \ref{fig_hardware}. The nonlinear receding horizon estimation (RHE) receives the measurements \eqref{eq_output} from onboard sensors, and estimates the full states and parameters \eqref{eq_state}-\eqref{eq_parameter}. The states and parameters estimated by the nonlinear RHE and a reference trajectory are used as inputs into the nonlinear RHC, which generates the control signal, i.e., the desired yaw rate. The desired yaw rate and speed as reference command signals are then sent to the Kangaroo x2 Motion Controller that adds self-tuning motion control to a Sabertooth dual 12A motor driver (Dimension Engineering, USA), which allows us to control the speed of the DC motors. The Kangaroo x2 Motion Controller, which is a two channel self-tuning PID controller, functions as the field robot's low-level controller by using feedback from the encoders attached to the DC motors to determine the required control signals. The Kangaroo x2 motion controller outputs the modified command signals to Sabertooth dual 12A motor driver, which correlates the given control signals to the necessary output voltages needed by the DC motors. In other words, the inner loop, the speed of the DC motors, is controlled by the Kangaroo x2 motion controller, which eliminates the need of manually tune the low-level controller, and executed at a rate of $50$-Hz.

\subsection{System Model with Traction Coefficients}

\begin{figure}[t!]
  \centering
  \includegraphics[width=3.2in]{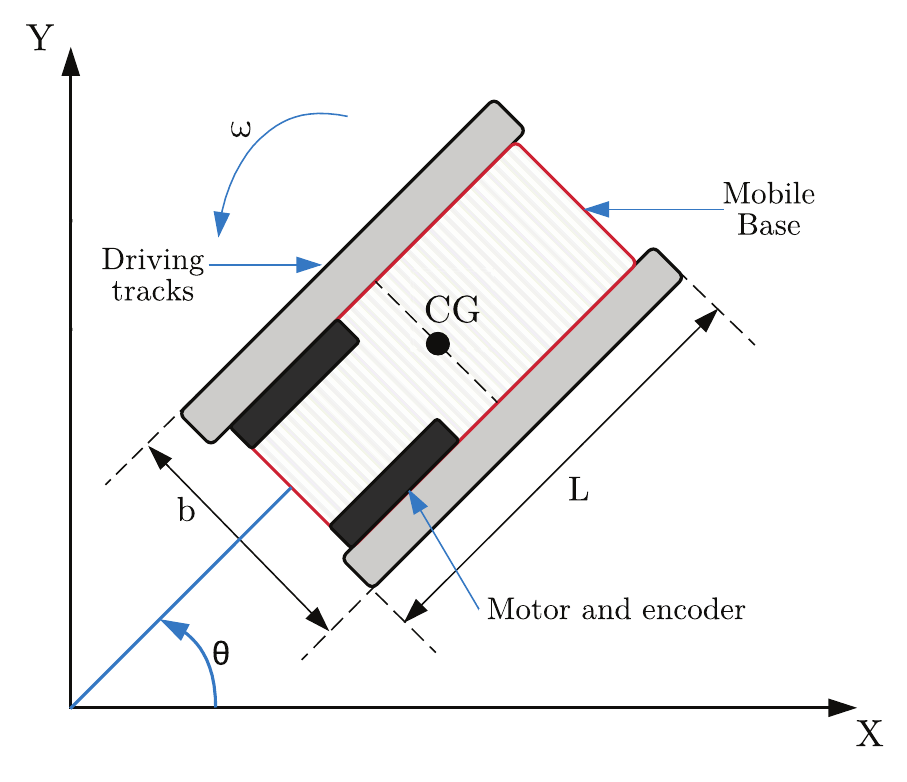}\\
  \caption{Schematic illustration of the field robot}\label{fig_mobilerobot}
\end{figure}

The schematic diagram of the field robot, which is a typical tracked mobile robot, is illustrated in Fig. \ref{fig_mobilerobot}. The velocities of two driven tracks result in linear velocity $\nu=( \nu_{l}+\nu_{r} )/2$ and angular velocity $\omega= (\nu_{r} - \nu_{l})/b$ with the distance between tracks $b$. The traditional kinematic model is formulated as follows \cite{Kanayama1990}:
\begin{eqnarray}\label{eq_tramodel}
			\dot{x} &=&  \nu \cos{\theta} \nonumber \\
        	\dot{y} &=&  \nu \sin{\theta} \nonumber \\
	\dot{\theta} &=&  \omega
\end{eqnarray}
where $x$ and $y$ denote the position of the field robot, $\theta$ denotes the yaw angle, $\nu$ denotes the speed, $\omega$ denotes the yaw rate.

The effect of terrain characteristics on robot performance and track should be well covered because the tracks are the unique connections between the ground and robot, and nearly all forces and moments applied to the robot are transmitted through the tracks \cite{1339393,4840546}. From track-terrain interaction dynamics, soil characteristic plays a vital role in determining vehicle speed and steering, which in turn are utilized for developing traction control algorithms. The knowledge of soil parameters of unknown terrain is then advantageous for improving vehicle performance \cite{7759528,7487413}. Therefore, instead of using the traditional kinematic model of a mobile robot \eqref{eq_tramodel}, an adaptive nonlinear kinematic model that is an extension of the traditional model is derived by adding two traction parameters ($\mu, \kappa$) to minimize deviations between the real-time system and system model of the field robot in this study. The field robot can be formulated with the following equations: 
\begin{eqnarray}\label{eq_systemmodel}
			\dot{x} &=& \mu \nu \cos{\theta} \nonumber \\
        	\dot{y} &=& \mu \nu \sin{\theta} \nonumber \\
	\dot{\theta} &=& \kappa \omega
\end{eqnarray}
where $\mu$ and $\kappa$ denote the traction parameters. It is noted that they must be between zero and one, and it is inherently arduous to measure them. If the traction parameters are equal to 1, i.e., $\mu=\kappa=1$, there exist no slip. The percentages of longitudinal and side slips are respectively found as $1-\mu$ and $1-\kappa$. These parameters show the effective speed $\mu v$ and steering $\kappa \omega$ of the field robot. To avoid bias, these traction parameters must be estimated along with the full system state in each iteration based on a number of past measurements.  

In the rest of the paper, we denote a nonlinear system model as
\begin{equation}\label{eq_nonlinearsystemmodel}
\dot{\xi}(t) = f \Big( \xi(t),u(t),p(t) \Big)
\end{equation} 
where $\xi$ $\in$ $\mathbb{R}^{n_{\xi}}$ is the state vector, $u$ $\in$ $\mathbb{R}^{n_{u}}$ is the control input, $p$ $\in$ $\mathbb{R}^{n_{p}}$ is the system parameter vector, $f(\cdot,\cdot,\cdot): \mathbb{R}^{n_{\xi}+n_{u}+n_{p}}  \longrightarrow \mathbb{R}^{n_{\xi}} $ is the continuously differentiable state update function and $f(0,0,p)=0 \; \forall t$. The derivative of $\xi$ with respect to $t$ is denoted by $\dot{\xi}$ $\in$ $\mathbb{R}^{n_{\xi}}$.
 
Similarly, a nonlinear measurement model denoted $z(t)$ can be described with the following equation:
\begin{equation}\label{eq_nonlinearmeasurementmodel}
z(t) = h \Big( \xi(t),u(t),p(t) \Big)
\end{equation}
where $h: R^{n_{\xi}+n_{u}+n_{p}} \longrightarrow R^{n_{z}}$ is the measurement function which describes the relation between the variables of the system model and the measured outputs of the real-time system. 

The state, parameters, input and output vectors are respectively denoted as follows:
\begin{eqnarray}
\label{eq_state}
\xi & = & \left[
  \begin{array}{ccc}
   x & y & \theta 
  \end{array}
  \right]^{T} \\ 
  \label{eq_parameter}
p & = & \left[
  \begin{array}{ccc}
   \nu &\mu & \kappa 
  \end{array}
  \right]^{T} \\ 
  \label{eq_input}
u & = & 
  \begin{array}{c}
   \omega 
  \end{array}
 \\ 
  \label{eq_output}
z & = & \left[
  \begin{array}{cccc}
   x & y & \nu & \omega 
  \end{array}
  \right]^{T}
\end{eqnarray}

\section{Nonlinear Receding Horizon Estimation and Control}\label{sec_rhec}
\subsection{Nonlinear Receding Horizon Estimation}

The least squares estimation problem uses all available past measurements to estimate the current values of the system states. In this estimation scheme, the number of measurements increases and the optimization problem is enlarged since a new measurement is acquired at every time instant. Therefore, solving the relevant optimization problem becomes infeasible in real-time as time approaches infinity. However, in practice, the computational burden of the estimation problem must remain constant so that the number of past measurements considered in the estimation scheme must also be constant. Furthermore, the relationship between the system model with current estimated parameters and measurements in the past can be inconsistent. To deal with this problem, past measurements are deemed less influential or left out. The idea of the RHE approach is to use only a constant number of the past measurements so that the oldest data are removed in the moving window when new data is acquired \cite{robertson1996a}. Thus, a fixed size window will be moving over the increasing measurement data so that the optimization problem will be fixed size. Then, the required method to incorporate the pre-information from the old measurements, which are not inside the moving window, and to prevent the heavy influence of the old measurements is discussed in subsection \ref{sec_thearrivalcost}. 

State estimation alone is not enough to know the system behavior in real-time when uncertain systems are considered; therefore, parameter estimation is required to determine uncertain parameters. A parametric least square estimation subject to the system model an/or boundary conditions was studied extensively in the literature, and there are many software packages to solve optimization problems for offline parameter estimation \cite{Houska2011a}. Moreover, two approximations have been proposed for online parameter estimation, which is necessary to determine system behavior in conjunction with state estimation accurately. In the first choice, model parameters are assumed as so-called random constant by a differential equation $\dot{\xi}_{p}=0$ with initial value $\xi_{p}(t_k)=p_{k}$. This approach results in time-invariant parameters over the estimation horizon. If jumps or drifts for parameters are expected, which is the case in practice under varying working conditions, a model bias will occur. To solve parameter jumps and drifts problem, the model parameters must be assumed as time-varying. Model parameters are assumed as so-called "random walk" by a differential equation $\dot{\xi}_{p}=\frac{d^{p}}{\triangle t}$ with sampling time $\triangle t$ and initial value $\xi_{p}(t_k)=p_{k}$ \cite{hughes1995}.

In the nonlinear RHE method, the parameters are assumed to be time-invariant in estimation horizon and not subject to process noise in the estimation horizon. However, it is also assumed that the parameters are time-varying Gaussian random variables in the arrival cost. For this reason, additional weighting factors must be added as the variance of the parameters noise and the parameters will appear only in the arrival cost. The extended weighting matrix $W$ $\in$ $\mathbb{R}^{(n_{xi}+n_{p})\times(n_{\xi}+n_{p})}$ is written as follows:
\begin{eqnarray}\label{Wupdate}
W = diag ( W_{\xi}, W_{p} )
\end{eqnarray}
where $W_{\xi}$ $\in$ $\mathbb{R}^{n_{\xi}\times n_{\xi}}$ and $W_{p}$ $\in$ $\mathbb{R}^{n_{p}\times n_{p}}$ denote the weighting matrices for the state noise and parameter pseudo-variance. Large values in $W$ result in large variations in the corresponding state and parameter estimates, and vice-versa. 

The nonlinear RHE is formulated as follows:
\begin{equation}
 \begin{aligned}
 & \underset{\xi(t),p,u(t)}{\text{min}}
 & &  \frac{1}{2} \bigg\{  \left\|
  \begin{array}{c}
    \hat{\xi}  - \xi (t_{k-N+1})  \\
    \hat{p} - p
 \end{array}
 \right\| ^{2}_{H_N} + \sum_{i=k-N+1}^{k}  \| z_m(t_i) - z (t_i) \|^{2}_{H_k} \bigg\}  \\
 & \text{ s. t. }
 &&  \dot{\xi}(t) = f \big(\xi(t),u(t),p \big) \\
 &&&  z(t) = h \big(\xi(t),u(t),p \big) \\
 &&& \xi_{min} \leq \xi(t) \leq \xi_{max} \\
 &&& p_{min} \leq p \leq p_{max} \qquad \qquad \forall t \in [t_{k-N},  t_k]\\
  \end{aligned}
    \label{mhe}
\end{equation}
where the deviations of the oldest states in the estimation horizon and the parameters from the prior estimates $\hat{\xi}$ and $\hat{p}$ are minimized by a symmetric positive semi-definite weighting matrix $H_N$. Deviations of the measured and system outputs in the estimation horizon are minimized by a symmetric positive semi-definite weighting matrix $H_k$, which is equal to the inverse of the measurement noise covariance matrix $V_{k}$ \cite{Ferreau}. As seen from the formulation, the objective function consists of two parts: the arrival and quadratic costs. The arrival cost stands for the early measurements $t=[t_{0,k-N+1}]$, and summarizes the deviations of the system outputs with respect to the measured outputs before the beginning of the estimation horizon. The quadratic cost stands for the recent measurements $t=[t_{t-N+1, k}]$, and corresponds to deviations of the system outputs with respect to the measured outputs within the estimation window. The estimation horizon is denoted by $N$, and lower and upper constraints on the model states and parameters are respectively denoted by $\xi_{min}$, $\xi_{max}$, $p_{min}$ and $p_{max}$.

\subsubsection{Arrival Cost}\label{sec_thearrivalcost}

The arrival cost in \eqref{mhe} is necessary for the stability of the nonlinear RHE. The estimation horizon for the arrival cost is equal to $t_{k-N}$ and approaches to infinity as time goes to infinity. Therefore, a real-time solution for the RHE formulation will be infeasible after a finite time. In order to forestall this problem, an iterative approach, a smoothed EKF update, is proposed for the approximation of the arrival cost. It is run with measurements at time $t_{k-N}$ and the inverted Kalman covariance matrix $H_{N}$ in every iteration to update Kalman estimates $\hat{\xi}, \hat{p}$. The main advantage of using the inverted Kalman covariance matrix is to have an adaptive weighting matrix. In case of using a constant weighting matrix, if the deviations in the arrival cost are small and the weighting matrix is large, then the chattering effect could be observed on the estimates. If the deviations in the arrival cost are large and the weighting matrix is small, then the estimates could not reach the true values in a reasonable time. 

The reference estimated values for the states and parameters $\hat{\xi}$ and $\hat{p}$ are acquired from the previous solution of RHE. Bounded arrival cost is a requirement because if the inverted Kalman covariance is too heavy, the arrival cost may approach infinity. Therefore, the contributions of the past measurements to the inverted Kalman covariance  $H_{N}$ are down-weighted by a process noise covariance matrix $W$ in \eqref{Wupdate}. It is to be noted that $H_{N}$ is upper bounded by the inverse of $W$ since $W^{-1} -H_{N}$ is positive semi-definite. The calculation of $H_N$ can be found in \cite{Robertson}.

\subsection{Nonlinear Receding Horizon Control}\label{section_RHC}

RHC approach has the ability to anticipate the system behavior on a finite horizon by minimizing a cost function consisting of the references, states, and inputs and deal with hard constraints on states and inputs. A nonlinear system model is described in \eqref{eq_nonlinearsystemmodel}. At every time step, the states and inputs have to fulfill the following:
\begin{equation}
\xi \in \mathbb{\Xi}, \;\;\; u \in \mathbb{U}
\end{equation}
where $\mathbb{U}$ is compact subset of $\mathbb{R}^{n_{u}}$, i.e., $\mathbb{U} \subseteq  \mathbb{R}^{n_{u}}$, $\mathbb{\Xi}$ is closed subset of $\mathbb{R}^{n_{\xi}}$, i.e.,  $\mathbb{\Xi} \subseteq  \mathbb{R}^{n_{\xi}}$, and each set contains the origin in its interior. The constraints on each input are uncoupled in the sense that the feasible regions of the inputs do not affect each other.

The following formulation is solved at each sampling time $t$:
\begin{equation}
 \begin{aligned}
 & \underset{\xi(t), u(t)} {\text{min}}
 & & \frac{1}{2} \bigg\{  \Big\{ \sum_{i=k+1}^{k+N-1}  \| \xi_{r} (t_{i}) - \xi (t_{i}) \|^{2}_{Q_{k}} + \| u_{r} (t_{i}) - u (t_{i}) \|^{2}_{R}  \Big\} \\
 &&& \qquad + \| \xi_{r} (t_{k+N}) - \xi (t_{k+N}) \|^{2}_{Q_{N}}  \bigg\} \\
 & \text{s. t.}
 && \xi(t_k) = \hat{\xi} (t_k) \\
 && & \dot{\xi}(t) = f \big(\xi(t), u(t), p \big) \\
 && & \xi_{min} \leq \xi(t) \leq \xi_{max} \quad t \in [t_{k+1}, t_{k+N}] \\
 && & u_{min} \leq u(t) \leq u_{max} \quad  t \in [t_{k+1}, t_{k+N-1}] \\
  \end{aligned}
  \label{NMPC}
\end{equation}
where $Q_{k} \in \mathbb{R}^{n_{\xi} \times n_{\xi}}$, $R \in \mathbb{R}^{n_{u} \times n_{u}}$ and $Q_{N} \in \mathbb{R}^{n_{\xi} \times n_{\xi}}$ are symmetric and positive semi-definite weighting matrices, $\xi_{r}$ and $u_{r}$ are the state and input references, $\xi$ and $u$ are the states and inputs, $t_k$ is the current time, $N$ is the prediction horizon, $\hat{\xi} (t_k)$ is the estimated state vector by the RHE, $\xi_{min}$, $\xi_{max}$, $u_{min}$ and $u_{max}$ denote respectively the upper and lower constraints on the state and input, $i=k+1, \hdots, k+N-1$, $u(t)=[u(t_{k+1}),\ldots,u(t_{k+N-1}]$ is the input sequence over prediction horizon $N$, and $\xi(t)$ is the state trajectory obtained by applying the control sequence $u(t)$ to the system. The first element of the input sequence $u(t)$ is applied to the systems:
\begin{equation}
u(t_{k+1},\xi_{k+1})= u^*(t_{k+1})
\label{U}
\end{equation}
and the aforementioned nonlinear RHC formulation is solved again over a shifted prediction horizon for the subsequent time instant. It is important to point out that the control input $u^*(t_{k+1})$ is precisely the same as it would be if all immeasurable states and parameters acquire values equal to their estimates based on the estimation up to current time $t_{k}$ due to the certainty equivalence principle.

The first cost is the stage cost, which is the cost throughout the prediction horizon, and the second cost is the terminal penalty, which is the cost at the end of the prediction horizon. The terminal penalty is stated for stability reasons \cite{Mayne,RawlingsMaynebook}.

The formulation in \eqref{NMPC} is a nonlinear and nonconvex optimization problem, and the computational complexity depends on the order of the system, the nonlinearity of the system, the horizon length and the nonlinear optimization solver.


\section{Implementations}\label{sec_implementations}
\subsection{Implementation of Nonlinear  RHE}

The inputs of the nonlinear RHE are the variables in the output vector \eqref{eq_output}, which includes the position, yaw rate and speed of the field robot. The outputs of the nonlinear RHE, the position, yaw angle, speed and traction parameters, are the full state and parameter vectors \eqref{eq_state}-\eqref{eq_parameter}. The nonlinear RHC requires full state and parameter as an input; therefore, the estimated values by the nonlinear RHE are fed to the nonlinear RHC to generate the desired yaw rate applied to the field robot.

The nonlinear RHE formulation is solved at every sampling instant with the following constraints on the traction parameters:
\begin{eqnarray}\label{eq_mhe_constraints}
0  \leq & \mu & \leq 1 \nonumber \\
0  \leq & \kappa & \leq 1
\end{eqnarray}

The weighting matrix  $H_{k}$ represents noise characteristics of sensors and is determined using standard deviation of measurements. The measurements have been perturbed by Gaussian noise with standard deviation of $\sigma_{x} = \sigma_{y} = 0.03$ m, $\sigma_{\omega} = 0.0175$ $rad/s$, $\sigma_{\nu} = 0.05$ $m/s$ based on experimental analysis. Additionally, the weighting matrix  $W$ represents process noise, which is chosen based on the objective. Low gain in the process noise results in better estimation accuracy; however, it causes time-lag between true and estimated values. Therefore, in this paper, the weighting coefficients for the measured states and parameters (e.g., x an y positions, speed) are selected large while the weighting coefficients for the immeasurable states and parameters (e.g., yaw angle and traction parameters) are selected small. The following weighting matrices $H_{k}$ and $W$ are used in nonlinear RHE:
\begin{eqnarray}\label{eq_mhe_weightingmatrices}
H_{k} = V_{k} ^{-1} & = &  diag(\sigma_{x}^{2},\sigma_{y}^{2},\sigma_{\nu}^{2}, \sigma_{\omega}^{2})^{-1} \nonumber \\
        & = & diag(0.03^{2},0.03^{2},0.05^{2},0.0175^{2})^{-1} \nonumber \\
W & = & diag(x^{2}, y^{2}, \theta^{2}, \nu^{2}, \mu^{2}, \kappa^{2}) \nonumber \\
      & = & diag(10.0^{2}, 10.0^{2}, 0.1^{2}, 1.0^{2}, 0.25^{2}, 0.25^{2})
\end{eqnarray}

\subsection{Implementation of Nonlinear RHC}

The nonlinear RHC formulation  is solved at every sampling instant with the following constraint on the input:
\begin{equation}\label{eq_mpc_constraints}
\num{-5.73}\si{\degree\per\second} \leq  \omega(t)  \leq \num{5.73}\si{\degree\per\second} 
\end{equation}

The state and input references for the field robot are changed online and defined as follows:
\begin{eqnarray}\label{eq_mpc_ref}
\xi_{r} = [x_{r},y_{r}, \theta_{r} ]^T \quad \textrm{and} \quad 
u_{r} = \omega_{r}
\end{eqnarray}
where $x_{r}$ and $y_{r}$ are the position references, $\omega_{r}$ is the yaw rate reference, and the yaw angle reference is calculated from the position references as follows:
\begin{eqnarray}\label{eq_mpc_ref_ya}
\theta_{r} = \atantwo{( \dot{y}_{r}, \dot{x}_{r})} + \lambda \pi
\end{eqnarray}
where $\lambda$ describes the desired direction of the field robot ( $\lambda=0$ for forward and $\lambda=1$ for backward). If the yaw rate reference calculated from the reference trajectory is used as the input reference, steady state error might occur in case of a mismatch between the system model and real system. Therefore, the measured yaw rate is used as the input reference to penalize the input rate in the objective function.

Three different weighting matrices $Q_{k}$ are selected to evaluate the yaw angle effect on the trajectory tracking in Section \ref{sec_scenarioA}.
\begin{eqnarray}
Q_{k}^{1} & = & diag(1,1,10)  \label{eq_mpc_weightingmatricesQ1}\\
Q_{k}^{2} & = & diag(1,1,0)  \label{eq_mpc_weightingmatricesQ2}\\
Q_{k}^{3} & = & diag(1,1,1) \label{eq_mpc_weightingmatricesQ3}
\end{eqnarray}
and the third one $Q_{k}^{3}$ is used in Section \ref{sec_scenarioB}. The weighting matrices $R$ and $Q_{N}$ for all experimental studies are selected as follows:
\begin{eqnarray}\label{eq_mpc_weightingmatricesRQN}
R = 10 \quad \textrm{and} \quad Q_{N}  = 10 \times Q_{k} 
\end{eqnarray}
The weighting matrix for the input $R$ is set larger than the one for the states $Q_{k}$ to ensure well damped closed-loop system behavior. The weighting matrix for the terminal penalty $Q_{N}$ is set to $10$ times larger than the one for the states $Q_{k}$. Therefore, the last deviations between the predicted states and their references in the prediction horizon are minimized in the objective function $10$ times more than the previous points in the prediction horizon. The reason is that the error at the end of the prediction horizon plays a very critical role for the stability issue of the control algorithm. 

\subsection{Extended Kalman Filter}

An extended Kalman filter (EKF) is designed based on the traditional kinematic model in \eqref{eq_tramodel}. The discrete-time model used by the EKF is written with a sampling interval $T_{s}$ as follows:
\begin{eqnarray}
\label{kinematicmodelEKF}
x_{k+1} & = & x_{k} + T_{s} \nu_{k} \cos{\theta_{k}} \nonumber \\
y_{k+1} & = & y_{k} + T_{s} \nu_{k} \sin{\theta_{k}} \nonumber \\
\theta_{k+1} & = & \theta_{k} + T_{s} \omega_{k}  
\end{eqnarray}

The general form of the estimated system model is:
\begin{eqnarray}
\label{generalmodelEKF}
\widehat{\xi}_{k+1} & = & f(\widehat{\xi}_{k}, u_{k}) + w_{k} \nonumber \\
\widehat{z}_{k+1} & = & h(\widehat{\xi}_{k}, u_{k}) + v_{k}
\end{eqnarray}
The differences between the kinematic model and the estimation model are the process noise $w_{k}$ and the observation noise $v_{k}$ both in the state and the measurement equations. They are both assumed to be independent with zero mean multivariate Gaussian noises with covariance matrices $W_{k}$ and $V_{k}$, respectively:
\begin{eqnarray}
\label{noise}
w_{k} \backsim N(0,W_{k}) \nonumber \\
v_{k} \backsim N(0,V_{k})
\end{eqnarray}
The covariance matrix $V_{k}$ is formulated in \eqref{eq_mhe_weightingmatrices}, and the covariance matrix $W_{k}$ is formulated as follows:
\begin{eqnarray}\label{eq_ekf_weightingmatrices}
W_{k} & = & diag(x^{2}, y^{2}, \theta^{2}, \nu^{2}) \nonumber \\
      & = & diag(10.0^{2}, 10.0^{2}, 0.1^{2}, 1.0^{2})
\end{eqnarray}
The convariance matrix $W_{k}$ for the EKF is different than the covariance matrix for the nonlinear RHE \eqref{eq_mhe_weightingmatrices} because the traction parameters are not estimated by the EKF. The EKF is designed based on the traditional kinematic model \eqref{eq_tramodel} because it cannot deal with constraints on parameters. If the traction parameters are estimated, their estimates might be lower than zero \cite{Haseltine2005}, which results in instability of the control algorithm.

\subsection{Solution Methods}

RHEC methods for systems require online solutions of nonlinear least square optimization problems at each sampling time. In this study, single solution method, consisting of a fusion between multiple shooting and generalized Gauss-Newton methods, has been used to solve both nonlinear RHEC optimization problems. This approach is valid because the formulation of nonlinear RHC problem is akin to that of nonlinear RHE problem. The generalized Gauss-Newton method, derived from the classical Newton method, was developed for least-squared problems. This method is advantageous because it does not require difficult computations of the second derivatives; however, it is challenging to foreknow the number of iterations to reach a solution of the desired accuracy. To overcome this challenge, the solution proposed in \cite{Diehl2,KRAUS2013135} has been used where the number of Gauss-Newton iterations is restricted to $1$, and the initial value of each optimization problem takes on the value of the previous one intelligently. Hence this improves the convergence of the Gauss-Newton method.

\begin{figure}[h!]
  \centering
  \includegraphics[width=3.4in]{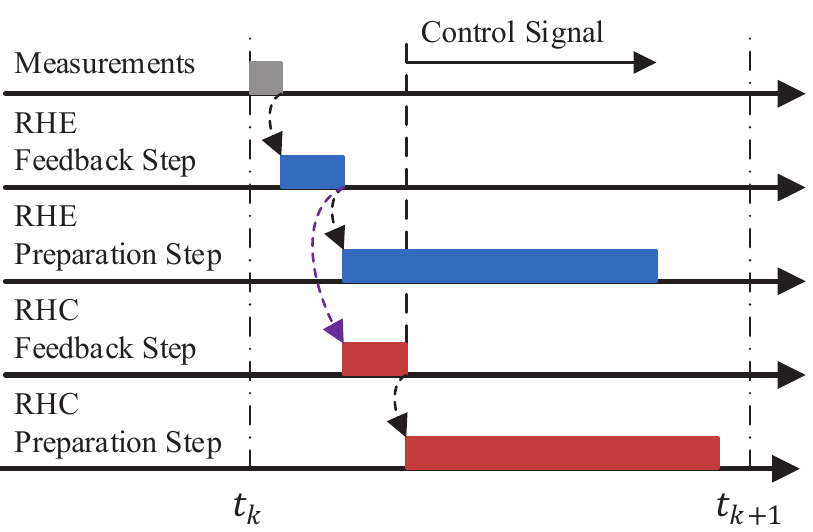}\\
  \caption{Illustration of the solution method}\label{fig_rhec_rti}
\end{figure}

Unlike \cite{VUKOV201564}, the Gauss-Newton iteration is divided into two parts: preparation and feedback parts. As illustrated in Fig. \ref{fig_rhec_rti}, the preparation part is executed prior to the feedback part, and the feedback part is executed after measurements for RHE and estimates for RHC are available. In the preparation part, the system dynamics are integrated with the previous solution, and objectives, constraints, and corresponding sensitives are evaluated. In the feedback part, a single quadratic programming is solved with the current measurements for the RHE and the current estimates for the RHC. Thus, the new estimates for the RHE and the new control signal for the RHC are obtained. Compared to the classical method, this method minimizes feedback delay and produces similar results with higher computational efficiency \cite{Diehl2}.

To solve the constrained nonlinear optimization problems in the nonlinear RHEC, the \emph{ACADO} code generation tool was used. \emph{ACADO} is an open source software package for optimization problems that generates C-code \cite{Houska2011a}. The obtained dense quadratic problem sub-problems are solved by the online quadratic problem solver \cite{Ferreau2008}. 


\section{Experimental Results}\label{sec_experimentalresults}

The aim of the real-time experiments is to track a predefined trajectory with a field robot exposed to varying soil conditions characterized by a wet and bumpy field. Highly accurate guidance is required to avoid crop, e.g., sorghum, damage during planting, weeding, and growth; however, slippage makes this guidance problem difficult. In this study, a space-based trajectory approach is employed rather than a time-based trajectory approach. The reason is that the speed of the field robot is featured with large uncertainty arising from the variation of terrain, load, turning resistance and surrounding obstacles. These uncertainties can impose large time deviations, which challenge the tracking of the reference time-based trajectory. However, it is to be pointed out that the field robot is not forced to be at a particular point on the trajectory at a certain sampling instant in space-based trajectory approach. 

The reference generation method in this paper is as follows. If the field robot starts off-track, the closest point on the target trajectory is computed first, and then the desired points in the horizon are determined. A trajectory consisting of straight and curved lines is tracked so that the performance of the designed framework for both scenarios is investigated. The prediction and estimation horizons for the nonlinear RHEC are set to 3 seconds.

\subsection{Scenario A: Effect of Yaw Angle Weight on RHC}\label{sec_scenarioA}

\begin{figure}[h!]
  \centering
  \includegraphics[width=3.6in]{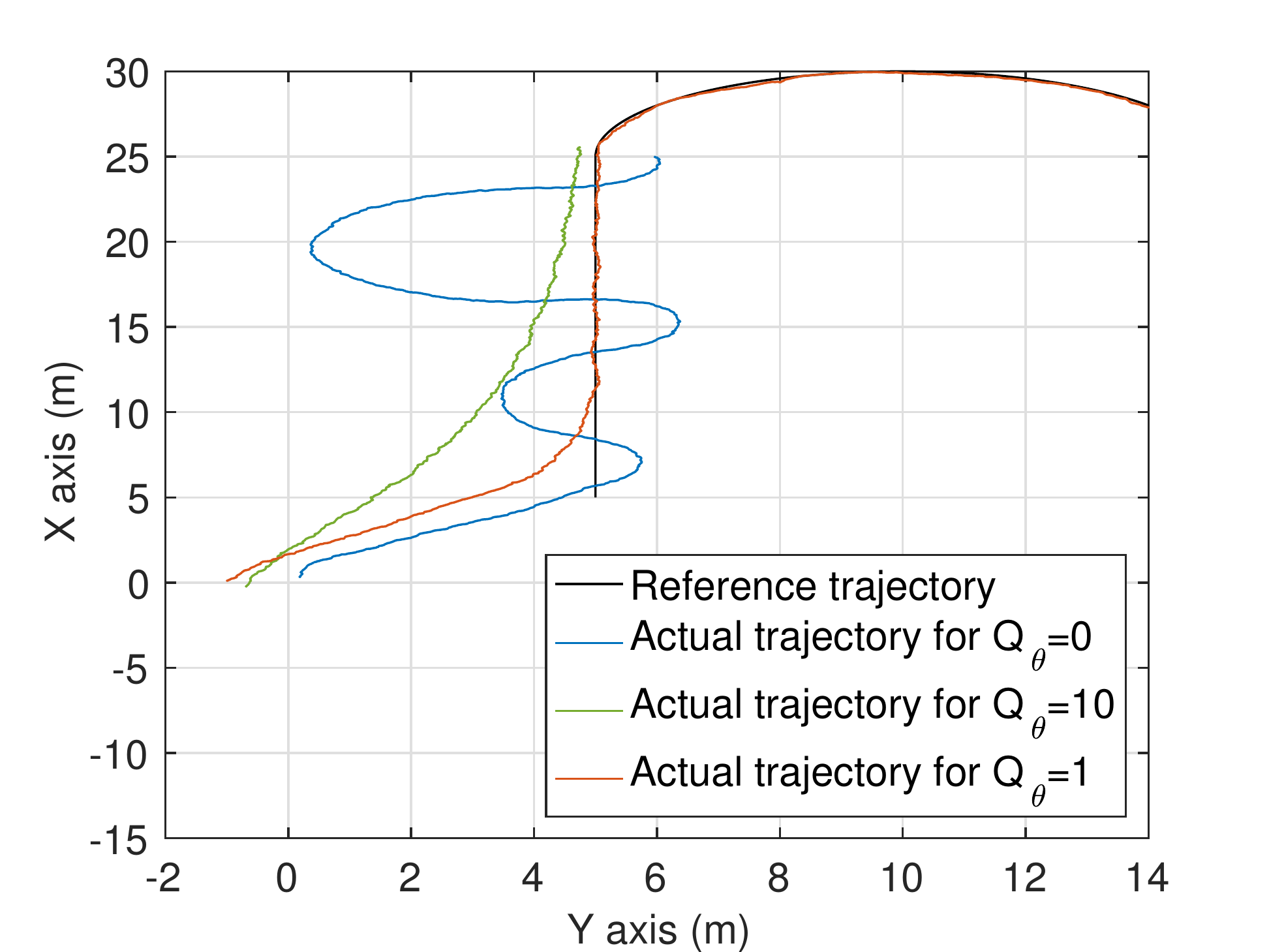}\\
  \caption{Reference and actual trajectories for different weighting of the yaw angle. These results clearly demonstrate that the penalization of yaw angle error is extremely important in RHC formulation. The presented approach in the paper should be contrasted with the typical approach of choosing small coefficients for R matrix, which essentially leads to low-gain control. By minimizing yaw angle error in the cost function in RHC formulation explicitly, large coefficients for R matrix can be employed, which leads to better tracking performance.  }\label{fig_mheQya}
\end{figure}

In this subsection, the coefficient of the yaw angle in the weighting matrix for nonlinear RHC is evaluated. Trajectory tracking performance of nonlinear RHCs based on nonlinear RHE is shown in Fig. \ref{fig_mheQya}. It is evident that if the coefficient of the yaw angle is larger than the coefficients for the position as in \eqref{eq_mpc_weightingmatricesQ1}, the field robot cannot reach the desired trajectory after a reasonable time. However, if the coefficient of the yaw angle is set to zero as in \eqref{eq_mpc_weightingmatricesQ2}, in other words, the yaw angle error between the reference yaw angle and actual yaw angle is not minimized, the field robot oscillates around the reference trajectory. To overcome this problem, the beginning of the desired points is defined as a fixed forward distance from the closest point on the reference trajectory at every sampling instant, and the optimum distance was selected regarding the speed of vehicles \cite{7525615}. 

In this paper, the minimization of the yaw angle error in the nonlinear RHC design is implemented to solve the aforementioned limitations. The coefficient of the yaw angle is set to equal to the coefficients for the position as in \eqref{eq_mpc_weightingmatricesQ3}. It can be seen from Fig. \ref{fig_mheQya} that the field robot reaches the reference trajectory after a reasonable time and stay on-track, which highlights the importance of correctly weighting of the yaw angle in the cost function in the space-based trajectory approach.

\subsection{Scenario B: Comparison of RHCs based on RHE and EKF}\label{sec_scenarioB}

\begin{figure}[h!]
  \centering
  \includegraphics[width=3.6in]{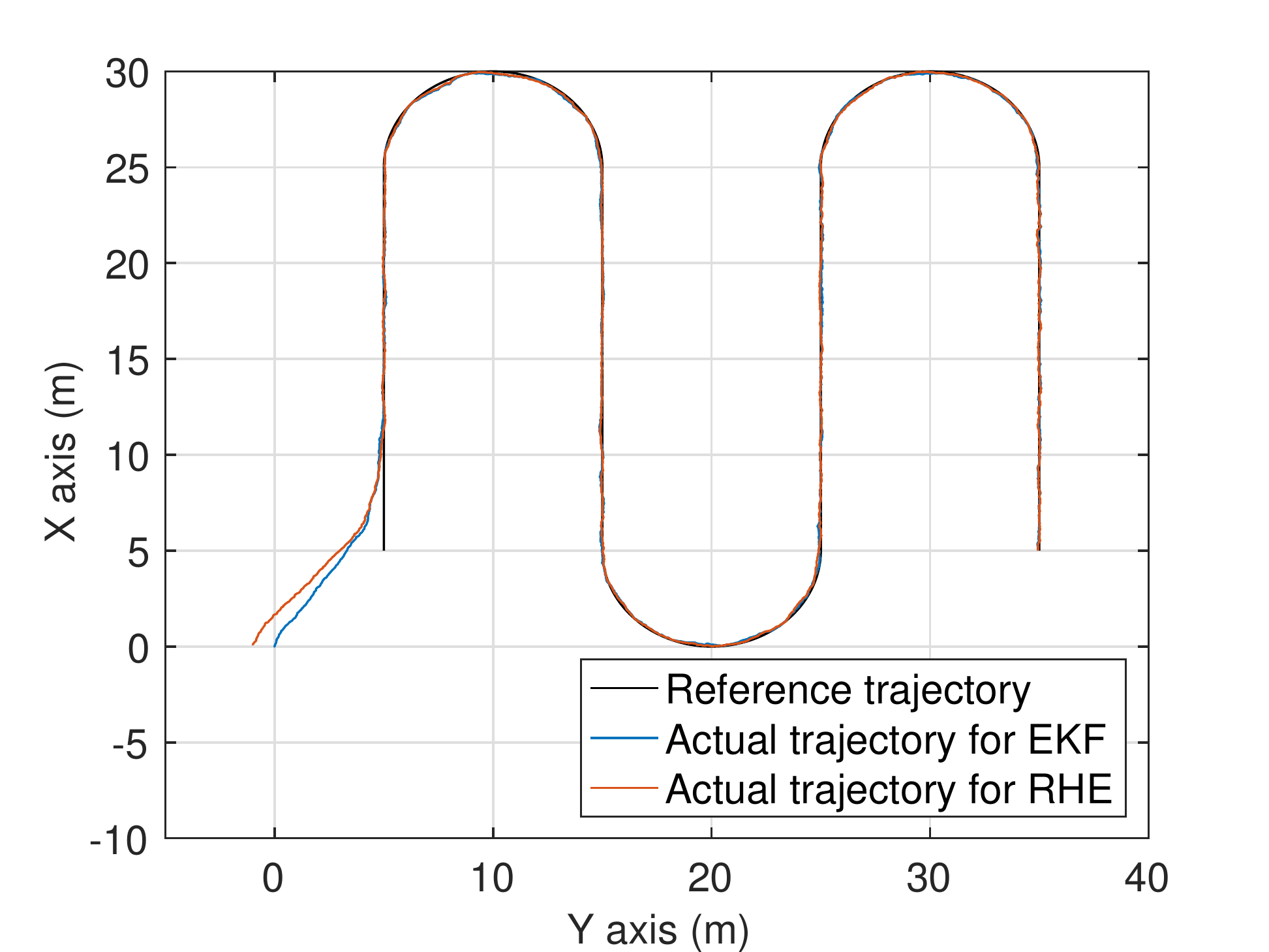}\\
  \caption{Reference and actual trajectories. The Euclidian errors are plotted in Fig. \ref{fig_error}. } \label{fig_Mheekf}
\end{figure}

Motivated by results from the previous section, the yaw angle error is minimized in the nonlinear RHC formulations based on the nonlinear RHE and the extended Kalman filter (EKF). As can be seen from Fig. \ref{fig_Mheekf}, the field robot reaches the reference trajectory after it is started off-track and stays-on track for nonlinear RHCs based on both the nonlinear RHE and the EKF.  

\begin{figure}[h!]
  \centering
  \includegraphics[width=6.5in]{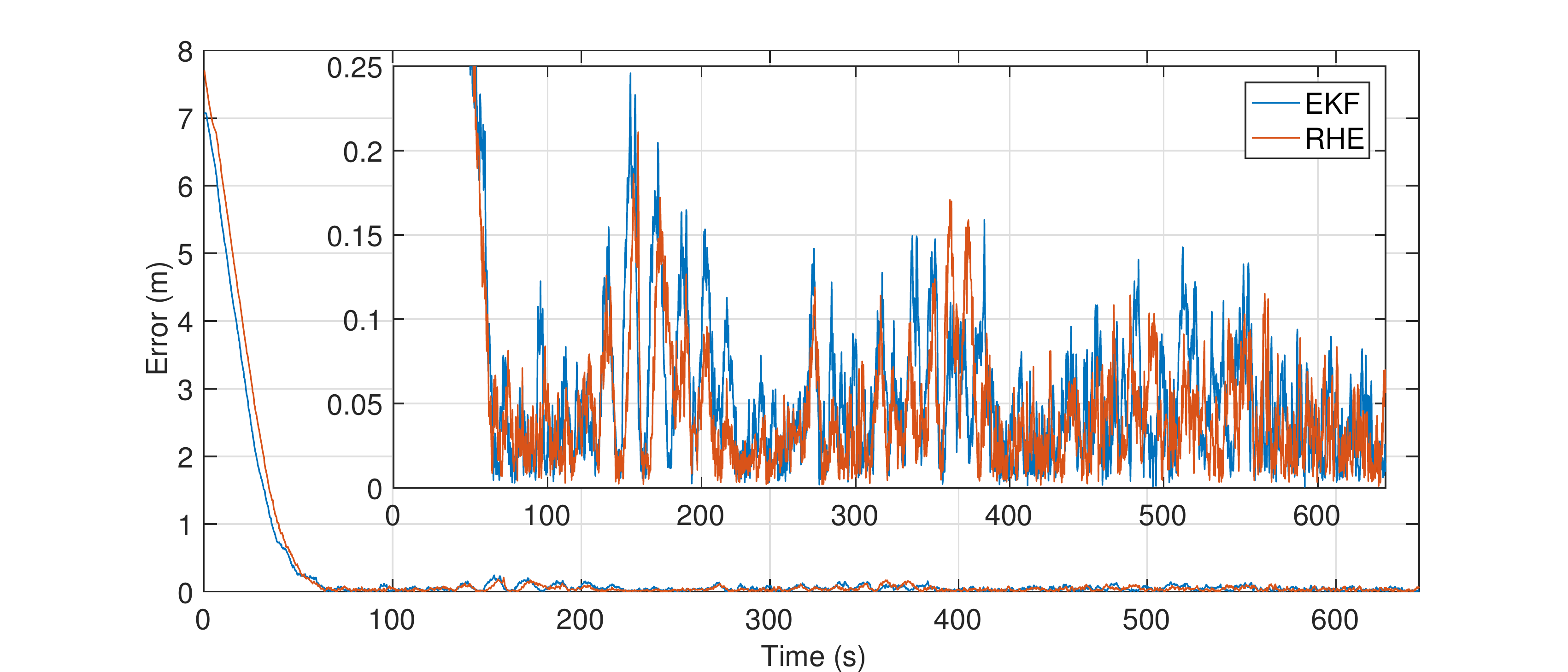}\\
  \caption{Euclidian error calculated using raw GNSS data}\label{fig_error}
\end{figure}

\begin{figure}[h!]
  \centering
  \includegraphics[width=3.8in]{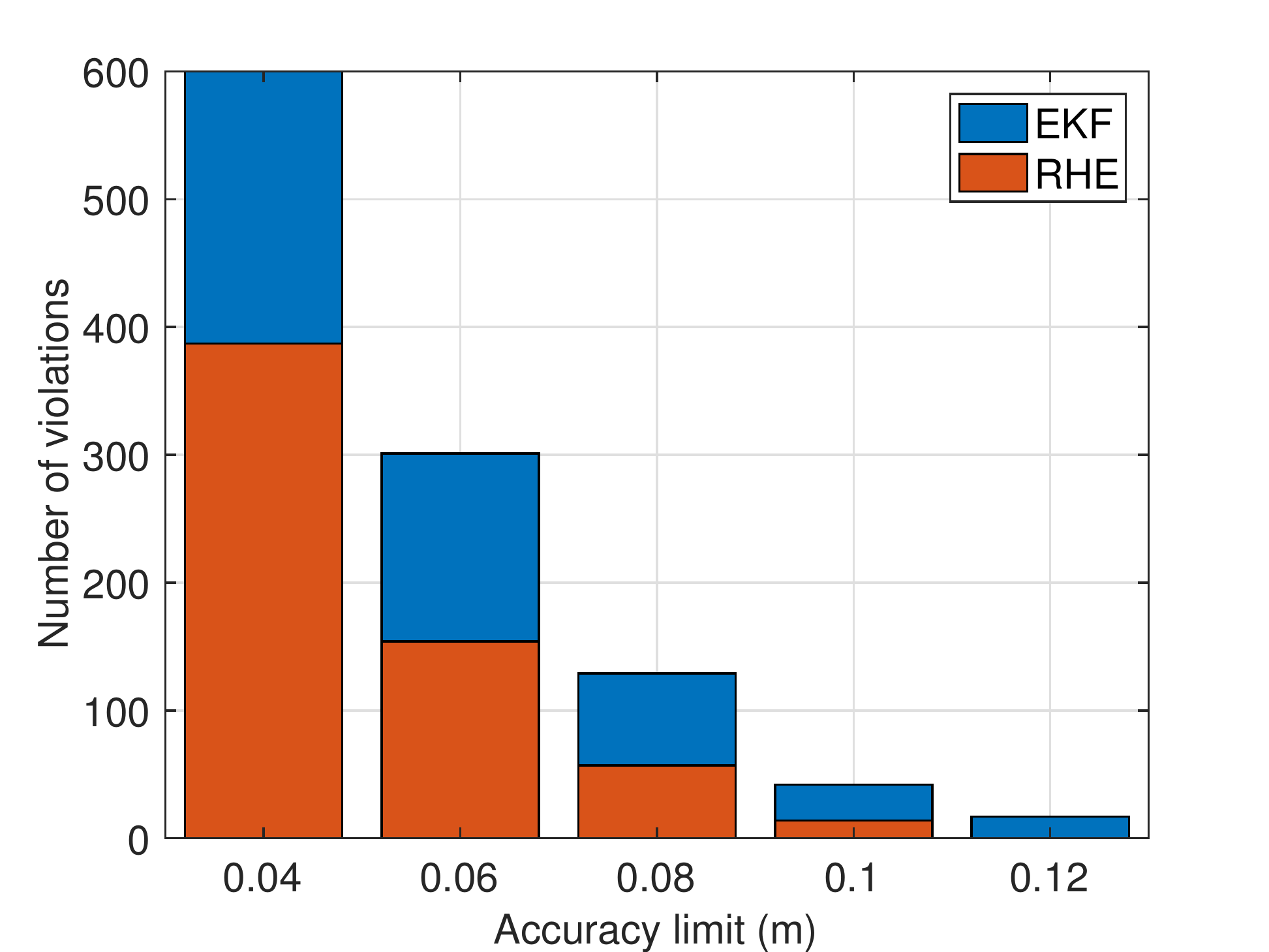}\\
  \caption{Number of violations for straight lines tracking. The available space on either side of the field robot is restricted to $0.12$ m. The field robot controlled by the RHEC framework does not violate this limit; therefore, it does not cause crop damage. }\label{fig_violations}
\end{figure}

The Euclidean errors calculated using raw GNSS data to the space-based reference trajectory for nonlinear RHCs based on both the nonlinear RHE and EKF are shown in Fig. \ref{fig_error}. The mean values of the Euclidian error for the nonlinear RHCs based on the nonlinear RHE and the EKF are $0.0423$ m and $0.0514$ m, respectively. The nonlinear RHEC framework benefits from traction parameter estimates and results in less error compared to the RHC based on the EKF. As previously mentioned in Section \ref{sec_intro}, the available space on either side of the robot is limited to only $0.12$ m; therefore, the error must be smaller than this limit to avoid crop damage and keep the robot-centered in the row. The number of violations is shown in Fig. \ref{fig_violations}. The results of multiple experiments indicate that the RHEC framework does not violate this error constraint, while the RHC based on the EKF violates it 17 times during the path tracking of straight lines. This demonstrates the capability of the RHEC framework.


\begin{figure}[h!]
  \centering
  \includegraphics[width=6.5in]{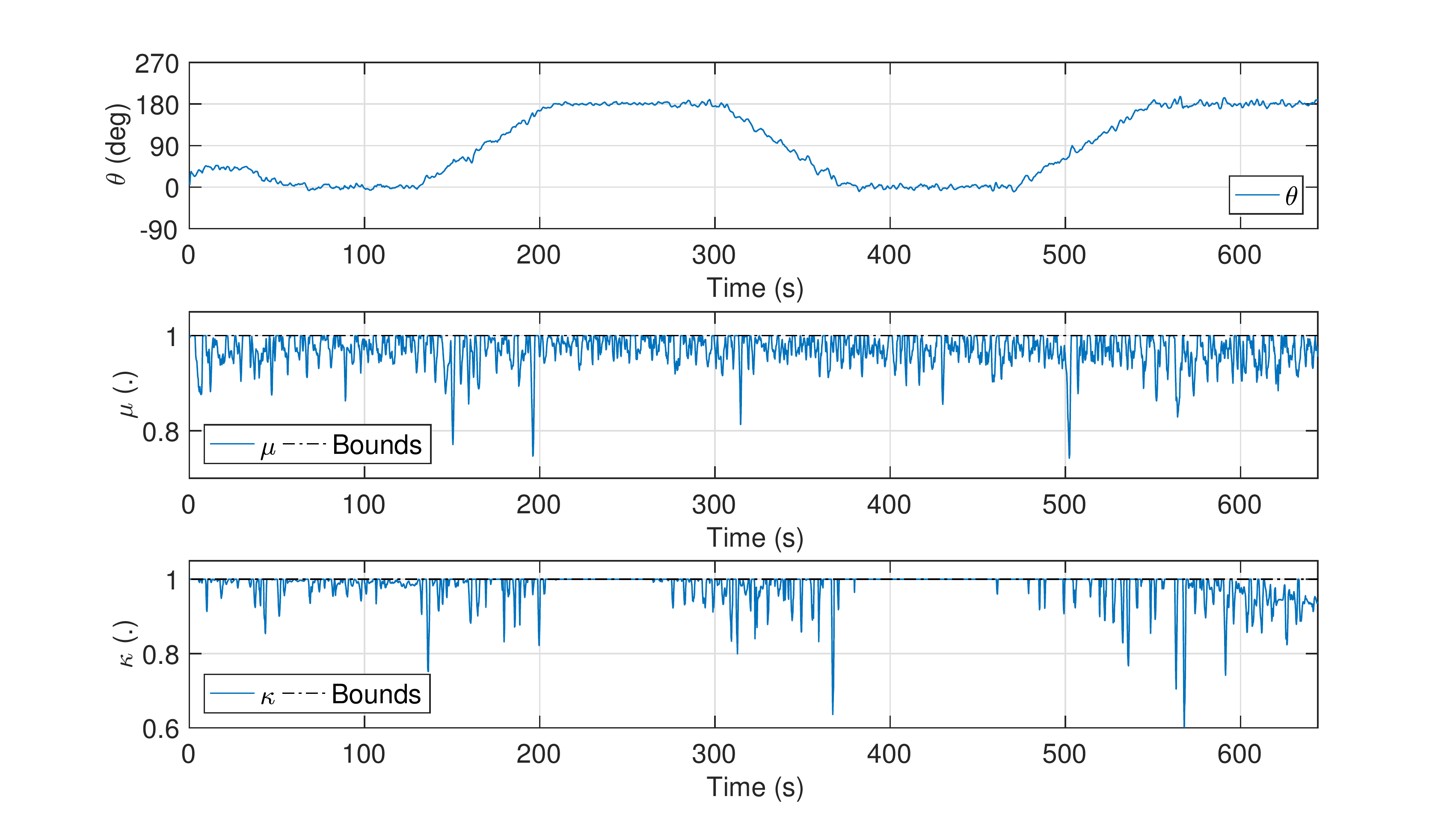}\\
  \caption{Nonlinear RHE output: The yaw angle and traction parameters estimates}\label{fig_estimates}
\end{figure}

The nonlinear RHE performance for the yaw angle and traction coefficients is shown in Fig. \ref{fig_estimates}. It is difficult to measure the yaw angle in practice, but it plays a vital role in the trajectory tracking problem. If it is not estimated, then the system model is deteriorated so that model-based controllers cannot be employed. As can be seen, the estimated values for the traction parameters are within the constraints specified in \eqref{eq_mhe_constraints} and consistently stay close to the upper bound. It is observed that model parameter estimates stabilize at certain values which secure stable trajectory-tracking. Therefore, these traction parameters' estimates play a significant role in the nonlinear RHC performance, and consequently the trajectory tracking performance. It is to be noted that hard-coding can result in more accurate trajectory tracking performance. However, the soil conditions must be in a steady-state, and the identification of parameters is required, which is a time-consuming task \cite{KAYACAN20141,6606388}. For this reason, the nonlinear RHE is beneficial because it provides the estimates of the traction parameters, and the framework ensures an accurate trajectory tracking. 

\begin{figure}[h!]
  \centering
  \includegraphics[width=6.5in]{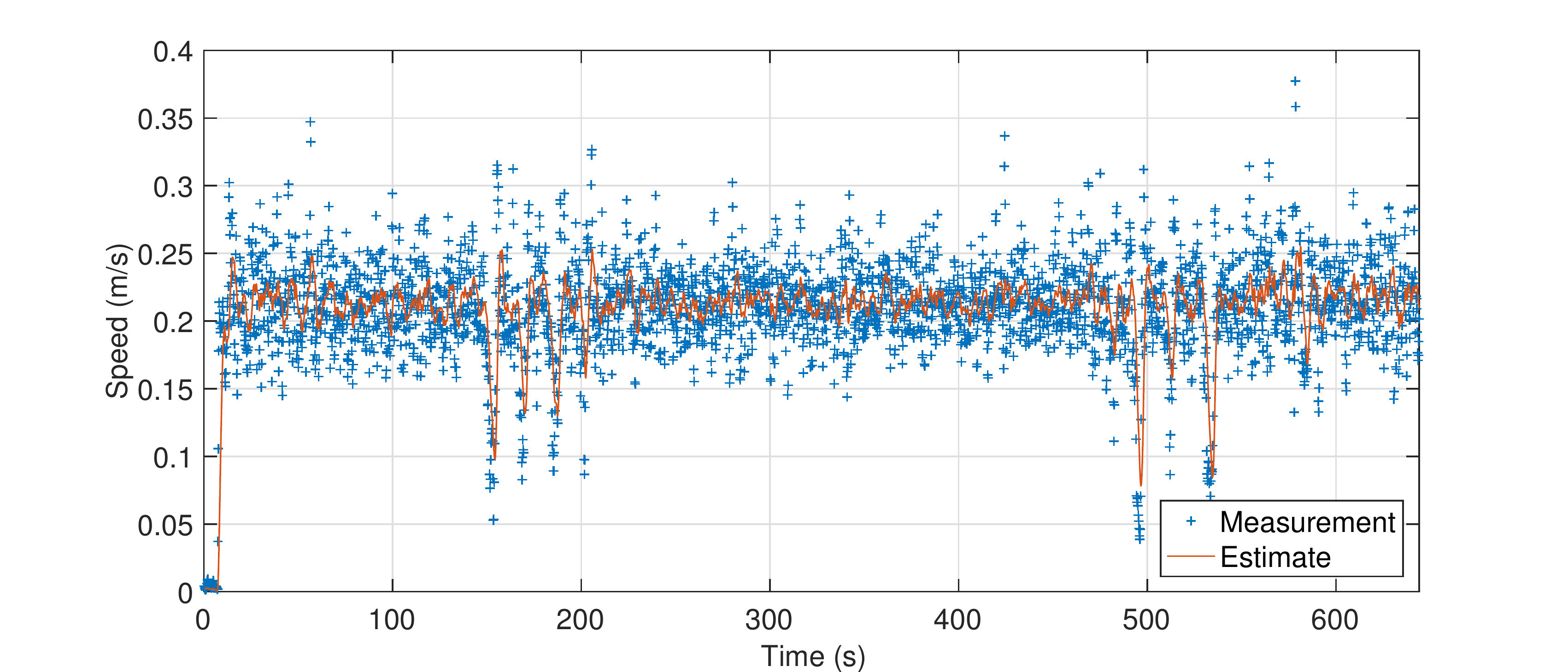}\\
  \caption{Actual and estimated speeds}\label{fig_speed}
\end{figure}

The measured and estimated speeds are shown in Fig. \ref{fig_speed}. The speed of the field robot is constant because a space-based trajectory approach is employed in the paper, and the nonlinear RHE can cope with noise on the speed measurements. 

\begin{figure}[h!]
  \centering
  \includegraphics[width=6.5in]{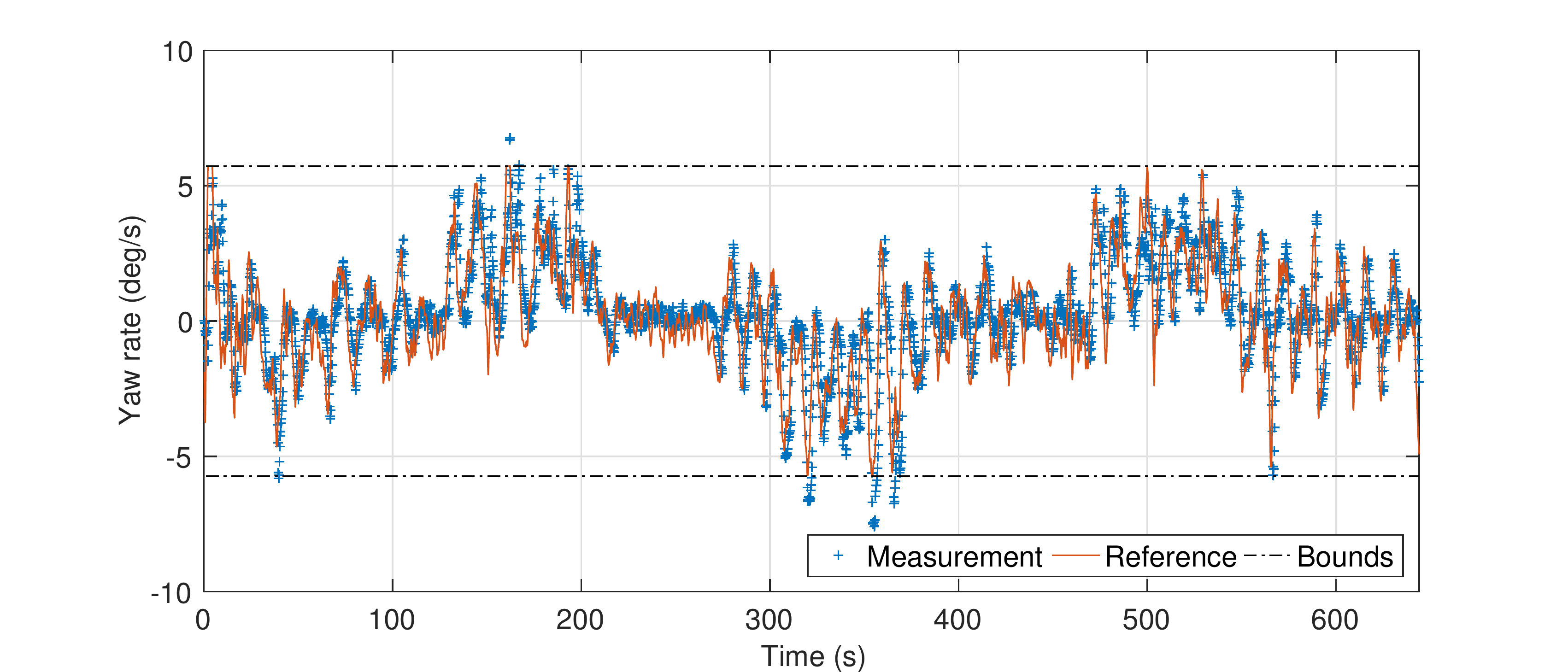}\\
  \caption{Reference and actual yaw rates}\label{fig_yawrate}
\end{figure}

The output of the nonlinear RHC, the reference yaw rate, is shown in Fig. \ref{fig_yawrate}. The output of the nonlinear RHC reaches the bounds, but never violates them. This demonstrates the capability of dealing with constraints on inputs. Moreover, the performance of the low-level controller for the speed of the DC motors is sufficient.

Real-time algorithms for RHE and RHC comprise respectively three states, two parameters, one control input and 15 prediction intervals, and three states, one control input, and 15 prediction intervals. The computations times for the nonlinear RHEC are summarized in Table \ref{tab_time}. As seen from the table, the average computation times for the nonlinear RHE and RHC are equal to $0.48$ ms and $0.40$ ms, respectively. Thus, the overall computation time for the nonlinear RHEC framework is equal to $0.88$ ms with a reasonable maximum computation time of $2.8534$ ms for real-time applications.

\begin{table}[h!]
\centering
\caption{Computation times of the nonlinear RHEC.}\label{tab_time}
\begin{tabular}{lccc}
  \hline 
   &  Minimum   &  Average   &  Maximum \\
   &   (ms)     &   (ms)     &  (ms)    \\
       \hline
  Nonlinear RHE   & && \\
  Preparation &  0.2921 & 0.3523 & 1.0075 \\
  Feedback    & 0.0766 & 0.1309 & 0.4925 \\
  Overall        & 0.3687 & 0.4832 & 1.5000 \\
  \hline
      \hline
Nonlinear RHC  & && \\
Preparation  & 0.2725 & 0.3751  & 1.2077 \\
Feedback     & 0.0149  & 0.0215 & 0.1457 \\
Overall         & 0.2874 & 0.3966  & 1.3534 \\
  \hline
\end{tabular}
\end{table}

\begin{figure}[h!]
  \centering
  \includegraphics[width=6.5in]{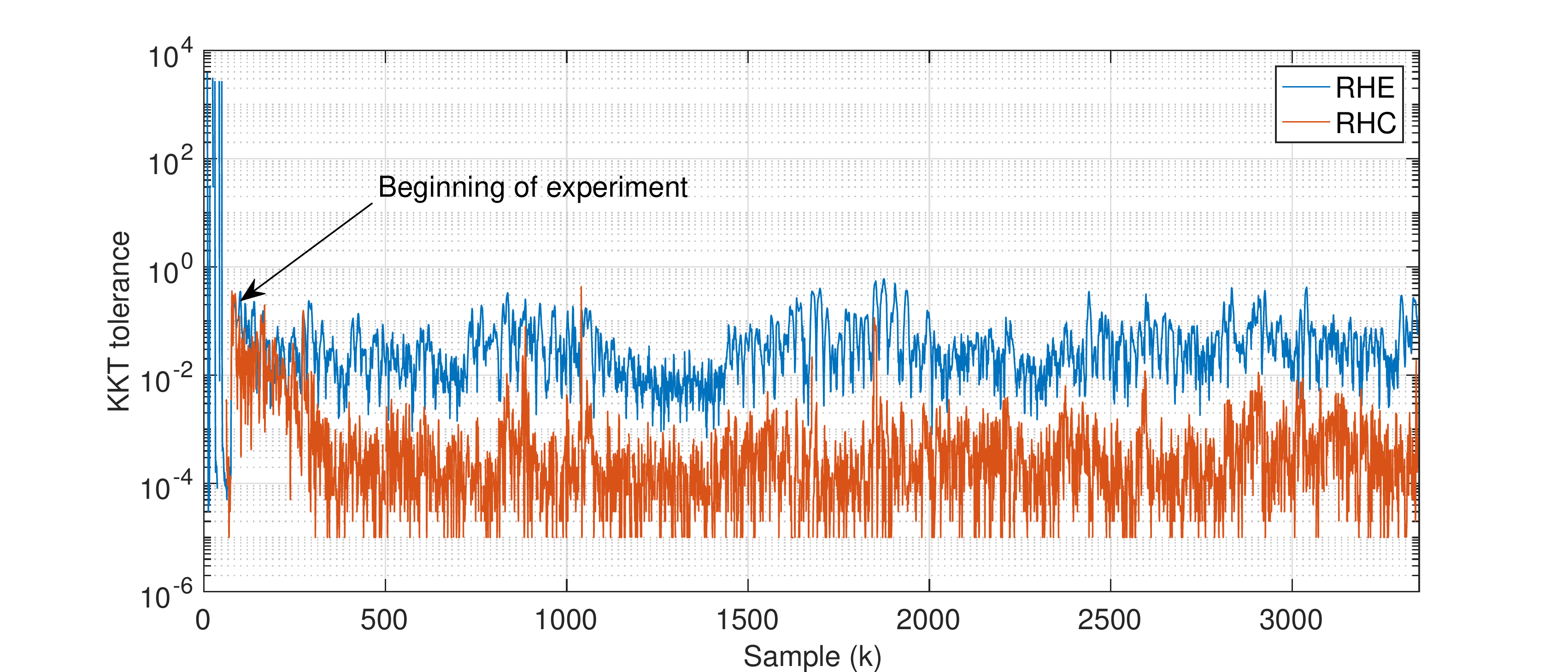}\\
  \caption{KKT tolerances}\label{fig_kkt}
\end{figure}

A single quadratic programming iteration at each time instant may result in a sub-optimal solution; therefore, it is necessary to observe the Karush-Kuhn-Tucker (KTT) tolerance, which measures the performance of RHEC methods in terms of optimality. The KKT tolerances for the RHEC are shown in Fig. \ref{fig_kkt}. It is to be noted that the KKT tolerance is equal to zero for linear systems due to the fact that a quadratic programming at each iteration is solved precisely. Therefore, non-zero values demonstrate the non-linearity of the RHEC problems. The largest KKT tolerances for the RHC arise at the beginning of the experiments because the field robot initially started off-track in the sense that error values are large. The reason is that since a single iteration is completed, the obtained solution is only a rough approximation to the optimal solution. It can be seen that the KKT tolerance decreases with time until the field robot stays on-track.  The trajectory references continuously change throughout the experiment; therefore, the KKT tolerance cannot further decrease. Furthermore, although the KKT tolerance for the RHE is large at the beginning due to poor initialization, it decreases after several time periods.

\section{Conclusion}\label{sec_conc}

A computationally efficient, nonlinear RHEC framework based on an adaptive nonlinear model has been investigated for the guidance of the field robot. The experimental results in the field trials have shown that the nonlinear RHE can perform precise online estimation of the immeasurable states and parameters. Also, the nonlinear RHC has the capability of accurately controlling the field, phenotyping robot with reasonable accuracy. The importance of correctly weighting of the yaw angle in the cost function is highlighted for the space-based trajectory approach. The mean value of the Euclidean error to the reference trajectory is $0.0423$ m. Additionally, the RHEC framework does not violate the upper bound of the required tracking accuracy, i.e., $0.12$ m, while the RHC based on the EKF violates it 17 times for the tracking of straight lines. Although RHEC framework is computationally intensive, the proposed solution method is computationally efficient, with a mean total required computation time of $0.88$ ms and the worst-case computation time of $2.85$ ms. 

The high accuracy of this nonlinear RHEC framework is of importance to plant breeders and farmers because there is a growing need for rapid, mobile, field-based phenotyping technologies that can navigate safely and autonomously through row based crops. By developing a framework that can handle unknown and immeasurable slip parameters, field robots will be able to navigate a variety of agricultural field conditions without damaging the crop.

Recent developments in microprocessors technology and fast solution tools for RHEC framework have changed the well-known paradigm in a way that the belief of using RHEC for only relatively slow dynamic systems is no longer true. The comparative results presented in this paper also show that RHEC implementations for agile robotic systems are eligible to obtain highly accurate tracking performance.


\subsubsection*{Acknowledgments}
The information, data, or work presented herein was funded in part by the Advanced Research Projects Agency-Energy (ARPA-E), U.S. Department of Energy, under Award Number DE-AR0000598. The views and opinions of authors expressed herein do not necessarily state or reflect those of the United States Government or any agency thereof.

Dr. Erkan Kayacan is a postdoctoral researcher at Massachusetts Institute of Technology since February 2018. 

\bibliographystyle{apalike}
\bibliography{HPCRefs}

\begin{thebibliography}{}

\bibitem[KRA, ]{KRAUS2013135}


\bibitem[Backman et~al., 2012]{Backman2012}
Backman, J., Oksanen, T., and Visala, A. (2012).
\newblock Navigation system for agricultural machines: Nonlinear model
  predictive path tracking.
\newblock {\em Computers and Electronics in Agriculture}, 82:32 -- 43.

\bibitem[Bekker, 1956]{Bekker1956}
Bekker, M. (1956).
\newblock {\em Theory of Land Locomotion}.
\newblock University of Michigan Press, Ann Arbor, MI.

\bibitem[Buckeridge et~al., 2014]{buckeridge2014plants}
Buckeridge, M.~S., Carpita, N.~C., McCann, M.~C., et~al. (2014).
\newblock {\em Plants and bioenergy}.
\newblock Springer,.

\bibitem[Diehl et~al., 2002]{Diehl2}
Diehl, M., Bock, H., Schlöder, J.~P., Findeisen, R., Nagy, Z., and
  Allg{\"{o}}wer, F. (2002).
\newblock Real-time optimization and nonlinear model predictive control of
  processes governed by differential-algebraic equations.
\newblock {\em Journal of Process Control}, 12(4):577 -- 585.

\bibitem[Falcone et~al., 2007]{Falcone2007}
Falcone, P., Borrelli, F., Asgari, J., Tseng, H.~E., and Hrovat, D. (2007).
\newblock Predictive active steering control for autonomous vehicle systems.
\newblock {\em Control Systems Technology, IEEE Transactions on},
  15(3):566--580.

\bibitem[Ferreau et~al., 2012]{Ferreau}
Ferreau, H., Kraus, T., Vukov, M., Saeys, W., and Diehl, M. (2012).
\newblock High-speed moving horizon estimation based on automatic code
  generation.
\newblock In {\em Decision and Control (CDC), 2012 IEEE 51st Annual Conference
  on}, pages 687--692.

\bibitem[Ferreau et~al., 2008]{Ferreau2008}
Ferreau, H.~J., Bock, H.~G., and Diehl, M. (2008).
\newblock An online active set strategy to overcome the limitations of explicit
  mpc.
\newblock {\em International Journal of Robust and Nonlinear Control},
  18(8):816--830.

\bibitem[Fiorani and Tuberosa, 2013]{Fiorani2013}
Fiorani, F. and Tuberosa, R. (2013).
\newblock Future scenarios for plant phenotyping.
\newblock {\em Annu Rev Plant Biol}, 64:267--291.

\bibitem[Fukao et~al., 2000]{880812}
Fukao, T., Nakagawa, H., and Adachi, N. (2000).
\newblock Adaptive tracking control of a nonholonomic mobile robot.
\newblock {\em IEEE Transactions on Robotics and Automation}, 16(5):609--615.

\bibitem[Gu and Hu, 2006]{Dongbing2006}
Gu, D. and Hu, H. (2006).
\newblock Receding horizon tracking control of wheeled mobile robots.
\newblock {\em Control Systems Technology, IEEE Transactions on},
  14(4):743--749.

\bibitem[Haseltine and Rawlings, 2005]{Haseltine2005}
Haseltine, E.~L. and Rawlings, J.~B. (2005).
\newblock Critical evaluation of extended kalman filtering and moving-horizon
  estimation.
\newblock {\em Industrial \& Engineering Chemistry Research}, 44(8):2451--2460.

\bibitem[Houska et~al., 2011]{Houska2011a}
Houska, B., Ferreau, H.~J., and Diehl, M. (2011).
\newblock Acado toolkit—an open-source framework for automatic control and
  dynamic optimization.
\newblock {\em Optimal Control Applications and Methods}, 32(3):298 -- 312.

\bibitem[Hughes, 1995]{hughes1995}
Hughes, B. (1995).
\newblock Random walks and random environments.
\newblock {\em Oxford}, 1.

\bibitem[Iagnemma et~al., 2004]{1339393}
Iagnemma, K., Kang, S., Shibly, H., and Dubowsky, S. (2004).
\newblock Online terrain parameter estimation for wheeled mobile robots with
  application to planetary rovers.
\newblock {\em IEEE Transactions on Robotics}, 20(5):921--927.

\bibitem[Kanayama et~al., 1990]{Kanayama1990}
Kanayama, Y., Kimura, Y., Miyazaki, F., and Noguchi, T. (1990).
\newblock A stable tracking control method for an autonomous mobile robot.
\newblock In {\em Proceedings., IEEE International Conference on Robotics and
  Automation}, pages 384--389.

\bibitem[Kayacan, 2017]{7934317}
Kayacan, E. (2017).
\newblock Multiobjective $h_{\infty }$ control for string stability of
  cooperative adaptive cruise control systems.
\newblock {\em IEEE Transactions on Intelligent Vehicles}, 2(1):52--61.

\bibitem[Kayacan et~al., 2012a]{kayacan2012}
Kayacan, E., Bayraktaroglu, Z.~Y., and Saeys, W. (2012a).
\newblock Modeling and control of a spherical rolling robot: a decoupled
  dynamics approach.
\newblock {\em Robotica}, 30(4):671–680.

\bibitem[Kayacan et~al., 2018]{Kayacan2018}
Kayacan, E., Kayacan, E., Chen, I.-M., Ramon, H., and Saeys, W. (2018).
\newblock {\em On the Comparison of Model-Based and Model-Free Controllers in
  Guidance, Navigation and Control of Agricultural Vehicles}, pages 49--73.
\newblock Springer International Publishing, Cham.

\bibitem[Kayacan et~al., 2015a]{6695753}
Kayacan, E., Kayacan, E., Ramon, H., Kaynak, O., and Saeys, W. (2015a).
\newblock Towards agrobots: Trajectory control of an autonomous tractor using
  type-2 fuzzy logic controllers.
\newblock {\em IEEE/ASME Transactions on Mechatronics}, 20(1):287--298.

\bibitem[Kayacan et~al., 2012b]{KAYACAN2012863}
Kayacan, E., Kayacan, E., Ramon, H., and Saeys, W. (2012b).
\newblock Velocity control of a spherical rolling robot using a grey-pid type
  fuzzy controller with an adaptive step size.
\newblock {\em IFAC Proceedings Volumes}, 45(22):863 -- 868.
\newblock 10th IFAC Symposium on Robot Control.

\bibitem[Kayacan et~al., 2013]{6606388}
Kayacan, E., Kayacan, E., Ramon, H., and Saeys, W. (2013).
\newblock Modeling and identification of the yaw dynamics of an autonomous
  tractor.
\newblock In {\em 2013 9th Asian Control Conference (ASCC)}, pages 1--6.

\bibitem[Kayacan et~al., 2014a]{KAYACAN2014926}
Kayacan, E., Kayacan, E., Ramon, H., and Saeys, W. (2014a).
\newblock Distributed nonlinear model predictive control of an autonomous
  tractor-trailer system.
\newblock {\em Mechatronics}, 24(8):926 -- 933.

\bibitem[Kayacan et~al., 2014b]{KAYACAN20141}
Kayacan, E., Kayacan, E., Ramon, H., and Saeys, W. (2014b).
\newblock Nonlinear modeling and identification of an autonomous
  tractor-trailer system.
\newblock {\em Computers and Electronics in Agriculture}, 106(Supplement C):1
  -- 10.

\bibitem[Kayacan et~al., 2015b]{6870427}
Kayacan, E., Kayacan, E., Ramon, H., and Saeys, W. (2015b).
\newblock Robust tube-based decentralized nonlinear model predictive control of
  an autonomous tractor-trailer system.
\newblock {\em IEEE/ASME Transactions on Mechatronics}, 20(1):447--456.

\bibitem[Kayacan et~al., 2015c]{KAYACAN201578}
Kayacan, E., Kayacan, E., Ramon, H., and Saeys, W. (2015c).
\newblock Towards agrobots: Identification of the yaw dynamics and trajectory
  tracking of an autonomous tractor.
\newblock {\em Computers and Electronics in Agriculture}, 115:78 -- 87.

\bibitem[Kayacan and Peschel, 2016]{KAYACAN2016265}
Kayacan, E. and Peschel, J. (2016).
\newblock Robust model predictive control of systems by modeling mismatched
  uncertainty.
\newblock {\em IFAC-PapersOnLine}, 49(18):265 -- 269.
\newblock 10th IFAC Symposium on Nonlinear Control Systems NOLCOS 2016.

\bibitem[Kayacan et~al., 2016a]{7525615}
Kayacan, E., Peschel, J.~M., and Kayacan, E. (2016a).
\newblock Centralized, decentralized and distributed nonlinear model predictive
  control of a tractor-trailer system: A comparative study.
\newblock In {\em 2016 American Control Conference (ACC)}, pages 4403--4408.

\bibitem[Kayacan et~al., 2016b]{7302059}
Kayacan, E., Ramon, H., and Saeys, W. (2016b).
\newblock Robust trajectory tracking error model-based predictive control for
  unmanned ground vehicles.
\newblock {\em IEEE/ASME Transactions on Mechatronics}, 21(2):806--814.

\bibitem[Klancar and Skrjanc, 2007]{Klancar2007}
Klancar, G. and Skrjanc, I. (2007).
\newblock Tracking-error model-based predictive control for mobile robots in
  real time.
\newblock {\em Robotics and Autonomous Systems}, 55(6):460 -- 469.

\bibitem[Lee et~al., 2016]{7487413}
Lee, S.~U., Gonzalez, R., and Iagnemma, K. (2016).
\newblock Robust sampling-based motion planning for autonomous tracked vehicles
  in deformable high slip terrain.
\newblock In {\em 2016 IEEE International Conference on Robotics and Automation
  (ICRA)}, pages 2569--2574.

\bibitem[Lee and Iagnemma, 2016]{7759528}
Lee, S.~U. and Iagnemma, K. (2016).
\newblock Robust motion planning methodology for autonomous tracked vehicles in
  rough environment using online slip estimation.
\newblock In {\em 2016 IEEE/RSJ International Conference on Intelligent Robots
  and Systems (IROS)}, pages 3589--3594.

\bibitem[Lee et~al., 2001]{Lee2001}
Lee, T.-C., Song, K.-T., Lee, C.-H., and Teng, C.-C. (2001).
\newblock Tracking control of unicycle-modeled mobile robots using a saturation
  feedback controller.
\newblock {\em Control Systems Technology, IEEE Transactions on},
  9(2):305--318.

\bibitem[Lins~Barreto et~al., 2014]{Barreto2014}
Lins~Barreto, J.~C., Scolari~Conceicao, A.~G., Dorea, C. E.~T., Martinez, L.,
  and De~Pieri, E.~R. (2014).
\newblock Design and implementation of model-predictive control with friction
  compensation on an omnidirectional mobile robot.
\newblock {\em IEEE/ASME Transactions on Mechatronics}, 19(2):467--476.

\bibitem[Mayne et~al., 2000]{Mayne}
Mayne, D., Rawlings, J., Rao, C., and Scokaert, P. (2000).
\newblock Constrained model predictive control: Stability and optimality.
\newblock {\em Automatica}, 36(6):789 -- 814.

\bibitem[Naveau et~al., 2017]{7384453}
Naveau, M., Kudruss, M., Stasse, O., Kirches, C., Mombaur, K., and Souères, P.
  (2017).
\newblock A reactive walking pattern generator based on nonlinear model
  predictive control.
\newblock {\em IEEE Robotics and Automation Letters}, 2(1):10--17.

\bibitem[Oh et~al., 2015]{7126158}
Oh, H., Kim, S., and Tsourdos, A. (2015).
\newblock Road-map assisted standoff tracking of moving ground vehicle using
  nonlinear model predictive control.
\newblock {\em IEEE Transactions on Aerospace and Electronic Systems},
  51(2):975--986.

\bibitem[Rawlings and Mayne, 2009]{RawlingsMaynebook}
Rawlings, J.~B. and Mayne, D.~Q. (2009).
\newblock {\em Model predictive control: Theory and design}.
\newblock Nob Hill Pub.

\bibitem[Ray, 2009]{4840546}
Ray, L.~E. (2009).
\newblock Estimation of terrain forces and parameters for rigid-wheeled
  vehicles.
\newblock {\em IEEE Transactions on Robotics}, 25(3):717--726.

\bibitem[Robertson, 1996]{Robertson}
Robertson, D. (1996).
\newblock {\em Development and Statistical Interpretation of Tools for
  Nonlinear Estimation}.
\newblock Auburn University.

\bibitem[Robertson et~al., 1996]{robertson1996a}
Robertson, D.~G., Lee, J.~H., and Rawlings, J.~B. (1996).
\newblock A moving horizon-based approach for least-squares estimation.
\newblock {\em AIChE Journal}, 42(8):2209--2224.

\bibitem[Tverberg, 2012]{Tverberg2012}
Tverberg, G. (2012).
\newblock Oil supply limits and the continuing financial crisis.
\newblock {\em Energy}, 37(1):27--34.

\bibitem[Utstumo et~al., 2015]{7125124}
Utstumo, T., Berge, T.~W., and Gravdahl, J.~T. (2015).
\newblock Non-linear model predictive control for constrained robot navigation
  in row crops.
\newblock In {\em 2015 IEEE International Conference on Industrial Technology
  (ICIT)}, pages 357--362.

\bibitem[Vega-Sanchez and Ronald, 2010]{Vega2010}
Vega-Sanchez, M. and Ronald, P. (2010).
\newblock Genetic and biotechnological approaches for biofuel crop improvement.
\newblock {\em Curr Opin Plant Biol}, 21(2):218--224.

\bibitem[Vukov et~al., 2015]{VUKOV201564}
Vukov, M., Gros, S., Horn, G., Frison, G., Geebelen, K., J{\o}rgensen, J.,
  Swevers, J., and Diehl, M. (2015).
\newblock Real-time nonlinear mpc and mhe for a large-scale mechatronic
  application.
\newblock {\em Control Engineering Practice}, 45:64 -- 78.

\bibitem[Wang et~al., 2016]{Wang2016}
Wang, Y., Fan, C., Hu, H., Li, Y., Sun, D., Wang, Y., and Peng, L. (2016).
\newblock Genetic modification of plant cell walls to enhance biomass yield and
  biofuel production in bioenergy crops.
\newblock {\em Biotechnology Advances}, 34(5):997 -- 1017.

\bibitem[Yano and Tuberosa, 2009]{Yano2009}
Yano, M. and Tuberosa, R. (2009).
\newblock Genome studies and molecular genetics-from sequence to crops:
  genomics comes of age.
\newblock {\em Curr Opin Plant Biol}, 12(2):103--106.

\end{thebibliography}

\end{document}